\pdfoutput=1
\documentclass[11pt]{article}

\usepackage[]{acl}

\usepackage{times}
\usepackage{latexsym}

\usepackage[T1]{fontenc}
\usepackage[utf8]{inputenc}

\usepackage{algorithm}
\usepackage{microtype}
\usepackage{algcompatible,amsmath}
\usepackage{graphicx}
\usepackage{url}
\usepackage{multirow}
\usepackage{comment}
\usepackage{amssymb}
\usepackage{pifont}
\usepackage{balance}
\usepackage{mathtools}
\usepackage{colortbl}
\usepackage{bbm}
\usepackage{enumitem}
\usepackage{textcomp}
\usepackage{array}
\usepackage{booktabs}
\usepackage{tabularx}
\usepackage{CJKutf8}
\usepackage{newfloat}
\usepackage{listings}
\usepackage{framed}
\usepackage{xcolor}
\usepackage{soul}
\usepackage{listings}
\usepackage{bbding}
\usepackage{makecell}

\definecolor{codegreen}{rgb}{0,0.6,0}
\definecolor{codegray}{rgb}{0.5,0.5,0.5}
\definecolor{codepurple}{rgb}{0.58,0,0.82}
\definecolor{backcolour}{rgb}{0.95,0.95,0.92}

\lstdefinestyle{mystyle}{
    backgroundcolor=\color{backcolour},   
    commentstyle=\color{codegreen},
    keywordstyle=\color{magenta},
    numberstyle=\tiny\color{codegray},
    stringstyle=\color{codepurple},
    basicstyle=\ttfamily\footnotesize,
    breakatwhitespace=false,         
    breaklines=true,                 
    captionpos=b,                    
    keepspaces=true,                 
    numbers=left,                    
    numbersep=5pt,                  
    showspaces=false,                
    showstringspaces=false,
    showtabs=false,                  
    tabsize=2
}

\definecolor{Seashell}{RGB}{0, 225, 0}
\definecolor{Seashelll}{RGB}{255, 215, 0}
\definecolor{Firebrick4}{RGB}{0, 0, 0}%文字颜色红一点的

\makeatletter
\newcommand{\thickhline}{%
    \noalign {\ifnum 0=`}\fi \hrule height 2pt
    \futurelet \reserved@a \@xhline
}

\floatstyle{ruled}
\newfloat{listing}{tb}{lst}{}
\floatname{listing}{Listing}

\newcommand{\squishlist}{
\begin{list}{$\bullet$}
{ \usecounter{Lcount}
\setlength{\itemsep}{0pt}
\setlength{\parsep}{0pt}
\setlength{\topsep}{0pt}
\setlength{\partopsep}{0pt}
\setlength{\leftmargin}{2em}
\setlength{\labelwidth}{1.5em}
\setlength{\labelsep}{0.5em} } }
\newcommand{\squishend}{
\end{list} }

\title{GDA: Generative Data Augmentation Techniques for\\ Relation Extraction Tasks}

\author{Xuming Hu$^1$, Aiwei Liu$^1$, Zeqi Tan$^2$, Xin Zhang$^3$, Chenwei Zhang$^{4\dagger}$,\\ \textbf{Irwin King$^5$, Philip S. Yu$^{1,6}$}\\
  $^1$Tsinghua University, $^2$Zhejiang University, $^3$Harbin Institute of Technology (Shenzhen),\\ $^4$Amazon,
  $^5$The Chinese University of Hong Kong, $^6$University of Illinois at Chicago\\
  $^1$\texttt{\{hxm19,liuaw20\}@mails.tsinghua.edu.cn}\\
  $^4$\texttt{cwzhang@amazon.com}\\
  }

\begin{document}
\maketitle
\begin{abstract}
Relation extraction (RE) tasks show promising performance in extracting relations from two entities mentioned in sentences, given sufficient annotations available during training. Such annotations would be labor-intensive to obtain in practice. Existing work adopts data augmentation techniques to generate pseudo-annotated sentences beyond limited annotations. These techniques neither preserve the semantic consistency of the original sentences when rule-based augmentations are adopted, nor preserve the syntax structure of sentences when expressing relations using seq2seq models, resulting in less diverse augmentations. In this work, we propose a dedicated augmentation technique for relational texts, named \texttt{GDA}, which uses two complementary modules to preserve both semantic consistency and syntax structures. We adopt a generative formulation and design a multi-tasking solution to achieve synergies. Furthermore, \texttt{GDA} adopts entity hints as the prior knowledge of the generative model to augment diverse sentences. Experimental results in three datasets under a low-resource setting showed that \texttt{GDA} could bring {\em 2.0\%} F1 improvements compared with no augmentation technique. Source code and data are available\footnote{\url{https://github.com/THU-BPM/GDA}\\\phantom{00} $^\dagger$Corresponding Author.}.

% could bring {\em 2.4\% } F1 improvements performances compared with no augmentation technique.\footnote{We will open source data and code upon acceptance.}
\end{abstract}
\section{Introduction}
\label{sec:intro}

Relation Extraction (RE) aims to extract semantic relations between two entities mentioned in sentences and transform massive corpora into triplets in the form of (subject, relation, object). 
% For example, when given a sentence ``\textit{Donald Trump}$_{subject}$ was born in the \textit{United States}$_{object}$.'', we could extract a relation \texttt{BORN IN} between subject and object entities. 
Neural relation extraction models show promising performance when high-quality annotated data is available \cite{zeng2017incorporating,zhang2017position,peng2020learning}. While in practice, human annotations would be labor-intensive and time-consuming to obtain and hard to scale up to a large number of relations \cite{hu2020selfore,hu2021semi,hu2021gradient,liu2022hierarchical}. This motivates us to solicit data augmentation techniques to generate pseudo annotations.

% Data augmentation techniques \cite{shorten2019survey} expand the annotated data by generating pseudo synthetic data to bootstrap the performance of neural networks, which is widely adopted in computer vision \cite{perez2017effectiveness,mounsaveng2021learning} and speech \cite{ko2017study}. However, compared with computer vision and speech which can easily adopt hand-crafted rules (such as rotation, cropping, flipping, etc.) to modify the original annotated data while keeping the visual information unchanged, language is more ambiguous and domain-specific. Therefore, it is difficult to apply data augmentation techniques to natural language processing (NLP) tasks. 

A classical effort devoted to data augmentation in NLP is adopting rule-based techniques, such as synonym replacement \cite{zhang2015character,cai2020data}, random deletion \cite{kobayashi2018contextual,wei2019eda}, random swap \cite{min2020syntactic} and dependency tree morphing \cite{csahin2018data}. However, these methods generate synthetic sentences without considering their semantic consistencies with the original sentence, and may twist semantics due to the neglection of syntactic structures. Other successful attempts on keeping the semantic consistency of the sentences are model-based techniques. The popular back translation method \cite{dong2017learning,yu2018qanet} generates synthetic parallel sentences using a translation model to translate monolingual sentences from the target language to the source language. However, it works exclusively on sentence-level tasks like text classification and translation, which is not designed to handle fine-grained semantics in entity-level tasks like relation extraction. 
\citet{bayer2022data} design a specific method for RE tasks by fine-tuning GPT-2 to generate sentences for specific relation types. However, it cannot be used in practice because the model generates less diverse sentences -- it includes similar entities and identical relational expressions under the same relation.

To keep the generated sentences diverse while semantically consistent with original sentences, we propose a relational text augmentation technique named \texttt{GDA}.
% which exploits semantic relational signals to generate semantically consistent pseudo sentences and learns to increase the diversity of generated sentences by predicting the original sentences with approximate patterns. 
As illustrated in Figure \ref{fig:overview}, we 
% \chenwei{``apply'' sounds weak, also consider moving this statement to the later paragraph after you talk about the core ideas} 
adopt the multi-task learning framework 
% to generate the reconstructed original sentence and pattern approximation target sentence simultaneously. 
% \chenwei{need to discuss how the two ideas are complementry with each other}
% The core idea is that we 
% and build two seq2seq models which are complementary with each other:
with one shared encoder and two decoders that are complementary with each other:
One decoder aims to predict the original sentence by restructuring words in the syntactic structure, which can maintain the semantics of the original sentence and ensure the model has the ability to generate semantically consistent target sentence. However, restructuring the syntactic structure of the original sentence inevitably breaks the coherence.
% \chenwei{<- wouldn't reordering harm consistency? e.g. do we consider ``from a to b'' to ``from b to a'' consistent?}
% Therefore, another model performs dependency parsing on the original sentence, and search the target sentence which preserves and approximates the syntax patterns of the original sentence.
Therefore, another decoder preserves and approximates the syntax patterns of the original sentence by generating the target sentence with a similar syntax structure drawn from the existing data.
% , so it generates a variation of relation expressions, although the augmented sentence may talk about totally different head/tail entities
% and adopts the ``shortest Levenshtein distance'' method to search the target sentence with approximate 
% \chenwei{try to be specific -- what kind of pattern? syntax pattern? you want sentence structure to be similar, right?->}
% syntax pattern between two entities under the same relation. 
This decoder can not only keep the target sentences coherent but more importantly, ensure that the model could maintain the original syntax pattern when generating pseudo sentences. Therefore, different patterns under the same relation can be preserved, instead of predicting the same syntax pattern due to relational inductive biases \cite{sun2021progressive}, thereby increasing the diversity of augmented sentences.
% \chenwei{inductive bias cannot be avoided but only alleviated -- and more imporantly, it is not clear what kind of unwanted inductive bias we are talking about here} 
% alleviating relational inductive bias and 
% increasing the diversity of generated sentence. 
We further adopt an entity in the 
% \chenwei{source?}
% target sentence as a hint to the input of that decoder, which can serve as prior knowledge to help  generate diverse sentences by taking different entity hints during inference.
target sentence as a hint to the input of that decoder, which can serve as prior knowledge to control the content of generated sentences. During inference, we could generate diverse sentences by taking a variety of different entity hints and origin sentences with various syntax patterns as input.
% , to help bootstrap the relation signals and derive entity-oriented
% \chenwei{controllable may raise concerns for text generation people in terms of investigating how controllable the model is in our experiments}controllable 
% sentence in the generation process. 
% Note that the two seq2seq models share the same encoder, trying to complement each other to strengthen the encoder module. 
To summarize, the main contributions of this work are as follows:
\definecolor{gree}{RGB}{0, 160, 0}
\begin{figure}[t!]
    \centering \includegraphics[width=0.99\linewidth]{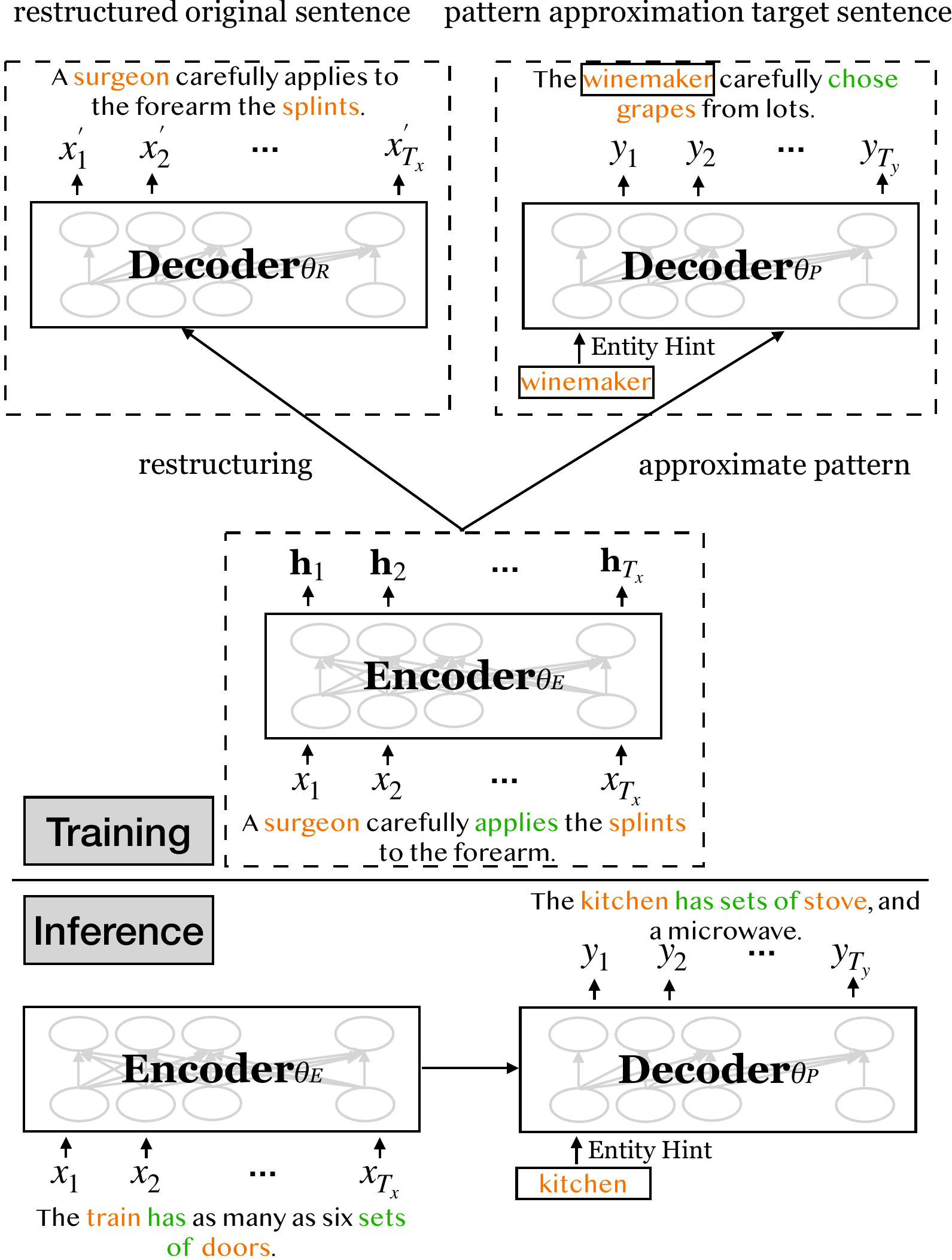}
    \caption{
    Overview of the proposed relational text augmentation technique with pattern approximation: \texttt{GDA}. We highlight the \textcolor{orange}{entities} and 
    \textcolor{gree}{pattern} in the sentences. We define the \textcolor{gree}{pattern} as the dependency parsing path between two entities.
    }
    \label{fig:overview}
    \vspace{-5mm}
\end{figure}

% The proposed technique has two steps, as shown in Figure \ref{fig:overview}, training a seq2seq generator leverages large pretrained language model T5 \cite{raffel2020exploring} to learn the generation of sentences with the same relation labels. This training formulation could not only utilize semantic relational signals to ensure the semantic consistency of generated sentences but leverage the generative model to generate coherent sentences. During the training process, we further add an entity in the target output sentence to the input sentence as a hint, which could help bootstrap the relation signals and derive entity-oriented controllable sentences in the generation process. After obtaining a trained generative model, we adopt the annotated sentences with sampled entity hints to generate coherent pseudo sentences with the same relation label. 

\squishlist
\item We study the task that focuses on the synergy between syntax and semantic preserving during data augmentation and propose a relational text augmentation technique \texttt{GDA}.
\item We adopt \texttt{GDA} which leverages the multi-task learning framework to generate semantically consistent, coherent, and diverse augmented sentences for RE task. Furthermore, entity hints from target sentences are served to guide the generation of diverse sentences.
\item We validate the effectiveness of \texttt{GDA} on three public RE datasets and low-resource RE settings compared to other competitive baselines.
\squishend

\section{Related Work}\label{related}

Data augmentation techniques have been widely used to improve the performance of models in the NLP tasks. The existing methods could be divided mainly into three categories: Rule-based techniques, Example interpolation techniques, and Model-based techniques \cite{feng2021survey}.
\paragraph{Rule-Based Techniques}
Rule-based techniques adopt simple transform methods. \citet{wei2019eda} proposes to manipulate some words in the original sentences such as random swap, insertion, and deletion. \citet{csahin2018data} proposes to swap or delete children of the same parent in the dependency tree, which could benefit the original sentence with case marking. \citet{chen2020finding} constructs a graph from the original sentence pair labels and augment sentences by directly inferring labels with the transitivity property.
\paragraph{Example Interpolation Techniques}
Example interpolation techniques such as MixText \cite{chen2020mixtext}, Ex2 \cite{lee2021neural}, and BackMix \cite{jin2022admix} aim to interpolate the embeddings and labels of two or more sentences. \citet{guo2020sequence} proposes SeqMix for sequence translation tasks in two forms: the hard selection method picks one of the two sequences at each binary mask position, while the soft selection method softly interpolates candidate sequences with a coefficient. Soft selection method also connects to existing techniques such as SwitchOut \cite{wang2018switchout} and word dropout \cite{sennrich2016improving}.
\paragraph{Model-Based Techniques}
Model-based techniques such as back translation \cite{sennrich2016improving}, which could be used to train a question answering model \cite{yu2018qanet} or transfer sequences from a high-resource language to a low-resource language \cite{xia2019generalized}. \citet{hou2018sequence} introduce a sequence to sequence model to generate diversely augmented data, which could improve the dialogue language understanding task. \citet{kobayashi2018contextual,gao2019soft} propose the contextualized word replacement method to augment sentences. \citet{anaby2020not,li2022data,bayer2022data} adopt language model which is conditioned on sentence-level tags to modify original sentences exclusively for classification tasks. Some techniques try to combine some simple data augmentation methods \cite{ratner2017learning,ren2021text} or add human-in-the-loop \cite{kaushik2019learning,kaushik2020explaining}. Other paraphrasing techniques \cite{kumar2020syntax,huang2021generating,gangal2021nareor} and rationale thinking methods \cite{hu2023think} also show the effectiveness of data augmentation methods. 

% \begin{table}
% \centering
% \scalebox{0.75}{
% \begin{tabular}{ccccc}
% \thickhline
% \multicolumn{2}{c}{\multirow{2}{*}{Methods}} & \multicolumn{3}{c}{Characteristics} 
% % & \multicolumn{4}{c}{Sina Weibo}  
% \\ \cmidrule(lr){3-5} 
% & &  Seman. & Coher. & Diver.  \\
% \midrule
% \multirow{2}{*}{\makecell[c]{Rule \\ Based}} & EDA & \Checkmark & \xmark & \Checkmark \\ 
% & Paraphrase Graph & \Checkmark & \xmark & \Checkmark \\ 
% \midrule
% \multirow{3}{*}{\makecell[c]{Example \\ Interpolation}} & MixText & \Checkmark & \xmark & \xmark \\ 
% & Ex2 & \Checkmark & \xmark & \xmark \\ 
% & BackMix & \Checkmark & \xmark & \xmark \\ 
% \midrule
% \multirow{5}{*}{\makecell[c]{Model \\ Based}} & Soft DA & \Checkmark & \xmark & \Checkmark \\ 
% & LAMBADA & \Checkmark & \Checkmark& \xmark \\ 
% & DARE & \Checkmark& \Checkmark & \xmark \\ 
% & Text Gen & \Checkmark & \Checkmark & \xmark \\ 
% & \textbf{\texttt{GDA} (Ours)} & \Checkmark & \Checkmark & \Checkmark\\ 

% \thickhline
% \end{tabular}}
% \caption{Characteristics comparison between different categories of techniques. ``Seman.'' means ``semantic consistency'', ``Coher.'' means ``coherence'', and ``Diver.'' means ``diversity''.}
% \label{tab:comparsion}
% \vspace{-4mm}
% \end{table}

\begin{table}
\centering
\scalebox{0.61}{
\begin{tabular}{clccc}
\thickhline
\multicolumn{2}{c}{\multirow{2}{*}{Methods}} & \multicolumn{3}{c}{Characteristics} 
% & \multicolumn{4}{c}{Sina Weibo}  
\\ \cmidrule(lr){3-5} 
& &  Seman. & Coher. & Diver.  \\
\midrule
\multirow{2}{*}{\makecell[c]{Rule \\ Based}} & \citet{wei2019eda} & \Checkmark & -- & \Checkmark \\ 
& \citet{chen2020finding} & \Checkmark & -- & \Checkmark \\ 
\midrule
\multirow{3}{*}{\makecell[c]{Example \\ Interpolation}} & \citet{chen2020mixtext} & \Checkmark & -- & --\\ 
& \citet{lee2021neural} & \Checkmark & -- & --\\ 
& \citet{jin2022admix} & \Checkmark & -- & -- \\ 
\midrule
\multirow{5}{*}{\makecell[c]{Model \\ Based}} & \citet{gao2019soft} & \Checkmark & -- & -- \\ 
& \citet{anaby2020not} & \Checkmark & \Checkmark& -- \\ 
& \citet{papanikolaou2020dare}& \Checkmark& \Checkmark & -- \\ 
& \citet{bayer2022data} & \Checkmark & \Checkmark & -- \\ 
& \textbf{\texttt{GDA} (Ours)} & \Checkmark & \Checkmark & \Checkmark\\ 

\thickhline
\end{tabular}}
\caption{Characteristics comparison between different categories of techniques. ``Seman.'' means ``semantic consistency'', ``Coher.'' means ``coherence'', and ``Diver.'' means ``diversity''.}
\label{tab:comparsion}
\vspace{-4mm}
\end{table}

\paragraph{Characteristics Comparsion}
We compare our \texttt{GDA} with other data augmentation techniques from the characteristics of semantic consistency, coherence, and diversity in Table \ref{tab:comparsion}. Note that the example interpolation techniques do not generate specific sentences, and only operates at the semantic embedding level. Therefore, we believe that they can only maintain semantic consistency. Compared with other SOTA data augmentation techniques, \texttt{GDA} uses a multi-task learning framework, which leverages two complementary seq2seq models to make the augmented sentences have semantic consistency, coherence, and diversity simultaneously.
% \vspace{-2mm}
\section{Proposed data augmentation technique}
The proposed data augmentation technique \texttt{GDA} consists of two steps: 1) Train a seq2seq generator. 2) Generate pseudo sentences. As illustrated in Figure \ref{fig:overview}, the first step adopts T5 \cite{raffel2020exploring} consisting of encoder and decoder parts as the seq2seq generator ($\theta$). The generator learns to convert two sentences with the same relation label. Specifically, the encoder part takes a sentence ${X = (x_{1}, x_{2}, ... , x_{T_{x}})}$ as input where named entities are recognized and marked in advance, and obtains the contextualized token embeddings ${H = (\textbf{h}_{1}, \textbf{h}_{2}, ... , \textbf{h}_{T_{x}})}$. The decoder part takes the ${H}$ as input and generates target sentence ${Y = (y_{1}, y_{2}, ... , y_{T_{y}})}$ word by word by maximizing the conditional probability distribution of ${p(y_{i}|y_{\textless i},H,\theta)}$. 
% Since both natural sentences are obtained from the annotated data, the augmented sentences generated by the trained generator will be more coherent and semantically consistent. 
% Besides, we additionally consider that the RE task is to obtain entity-level contextualized relational signals and perform relation classification, so we add the entity from the target output sentence to the input sentence as a hint for the generation process, which makes the generator pay more attention to entity-level relational signal reconstruction during training. 
The second step randomly selects an annotated sentence as input, and leverages the trained generator to generate pseudo sentence with entity marker and same relation label. Now, we introduce the details of each step.
% \vspace{-1mm}
\subsection{Train a seq2seq generator}
 
Training a seq2seq generator aims to obtain a generator that could augment annotated sentences to diverse, semantically consistent, and coherent pseudo sentences. In addition, the entities in the augmented pseudo sentences also need to be marked for entity-level relation extraction task. To achieve this goal, the generator must convert two sentences with the same relation label and emphasize contextualized relational signals at the entity level during the generation process. In practice, for each annotated sentence ${X = (x_{1}, x_{2}, ... , x_{T_{x}})}$, we adopt the labeling scheme introduced in \citet{soares2019matching}, and augment ${X}$ with four reserved tokens: ${[E_{sub}]}$, ${[/E_{sub}]}$, ${[E_{obj}]}$, ${[/E_{obj}]}$ to represent the start and end position of subject and object named entities respectively, and inject them to $X$. For example, ``A ${[E_{sub}]}$ surgeon ${[/E_{sub}]}$ carefully applies the ${[E_{obj}]}$ splints ${[/E_{obj}]}$ to the forearm.''. Then we feed the updated ${X}$ into the T5 encoder part to obtain contextualized token embeddings ${H}$: ${H=\operatorname{Encoder}(X)}$.

A natural paradigm for the decoder part to generate the target sentence is to select another sentence in the training set that has the same relation as the input sentence. \citet{bayer2022data} fine-tuned GPT-2 to generate sentences for specific relation types. However, it requires too much computational cost to train multiple GPT-2s for each relation type,
% chenwei{from our observation or from them?->}
and we observed no promising results are obtained. We attribute the reason to two aspects:
% 1) the ability of the model to generate semantically consistent sentences is not enhanced,
1) the diversity of generated sentences is not emphasized, resulting in the generation of sentences with similar patterns, and 2) the entity-level relation extraction task is not considered, resulting in missing entity information. 

In this paper, we propose to leverage the multi-task learning framework to address the above shortcomings, which performs two tasks: original sentence restructuring and original sentence pattern approximation. In practice, our framework consists of two seq2seq models that share the same encoder part, but employ two decoder parts to complete the two tasks, respectively.

\paragraph{Original sentence restructuring.}

The original sentence restructuring task aims to improve the ability of the model to generate semantically consistent sentences. As illustrated in Figure \ref{fig:overview}, the target generated sentence is just the restructured original sentence ${X^{'} = (x_{1}^{'}, x_{2}^{'}, ... , x_{T_{x}}^{'})}$ that has the same length and words as the original sentence. We adopt the pre-ordering rules proposed by \citet{wang2007chinese} in machine translation. These rules could modify the syntactic parse tree obtained from the original sentence and permutate the words by modifying the parsed tree. The target sentence is closer to the expression order of words without changing the semantics of the original sentence 
% \chenwei{<-important to highlight this nice feature in abstract/intro as it is definitely possible for readers to wrongly assume we use random shuffling, like in the bart paper. i might miss details in the wang's paper but after i read thru the whole manuscript, i have stronger feeling that ``reordering'' can be quite misleading -- the model does some ``re-structuring'' based on syntax, or syntax-based translations, to be more precise in my opinion}. 
Furthermore, since the entities are not changed, it is easy to mark the position of the entities, e.g.:
\begin{figure}[h!]
\vspace{-2mm}
    \centering
    \includegraphics[width=0.99\linewidth]{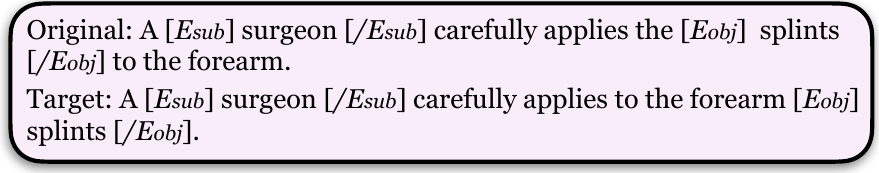}
\vspace{-4mm}
\end{figure}

In this way, the decoder network is employed to predict the restructured original sentence by maximizing ${p(X^{'}|H,\theta_{R})}$:
\begin{small}
\begin{equation}
\mathcal{L}_{\theta_{R}}=\sum_{m=1}^{M}\sum_{i=1}^{T_{x}} \log p\left(X_{i}^{'(m)} \mid X_{\textless i}^{'(m)},H^{(m)}, \theta_{R}\right),
\end{equation}
\end{small}
where ${\theta_{R}}$ denotes the parameters of the decoder part for original sentence restructuring. ${M}$ is the number of the training data.

\paragraph{Original sentence pattern approximation.}
The restructured sentence inevitably breaks the coherence of the original sentence. Therefore, we employ another seq2seq model to auxiliary predict unmodified sentences.
When we directly adopt seq2seq model for sentence generation \cite{bayer2022data}, The seq2seq model usually generates stereotyped sentences with the same pattern \cite{battaglia2018relational}.
% , called 
% \chenwei{``relational inductive bias'' seems more appropriate? inductive bias is a fairly generic ML term} 
% relational inductive bias. 
For example, for relation \texttt{Component-Whole}, using generative methods such as T5 will always tend to generate pattern ``\textit{consist of}'' with high frequency, rather than rare ``\textit{is opened by}'' and ``\textit{has sets of}'', which will limit the performance improvement of augmented sentences.
% \chenwei{<-sounds to me more like relation expression diversity, or predicate diversity. when you say inductive bias people would usually expect you to dicuss whether such bias is helpful or not to the task. also are those appropriate examples (appropriate as judged by the lev distance)?}. 
In this paper, we introduce the original sentence pattern approximation task to force the original and target sentences to have approximate patterns, so that the augmented sentences could maintain the pattern of the original sentences and enhance the sentence diversity. In practice, we define the \textbf{pattern} as the dependency parsing path between two entities. We assume this parsing path is sufficient to express relational signals \cite{peng2020learning}.
\begin{figure}[t!]
    \centering
    \includegraphics[width=0.98\linewidth]{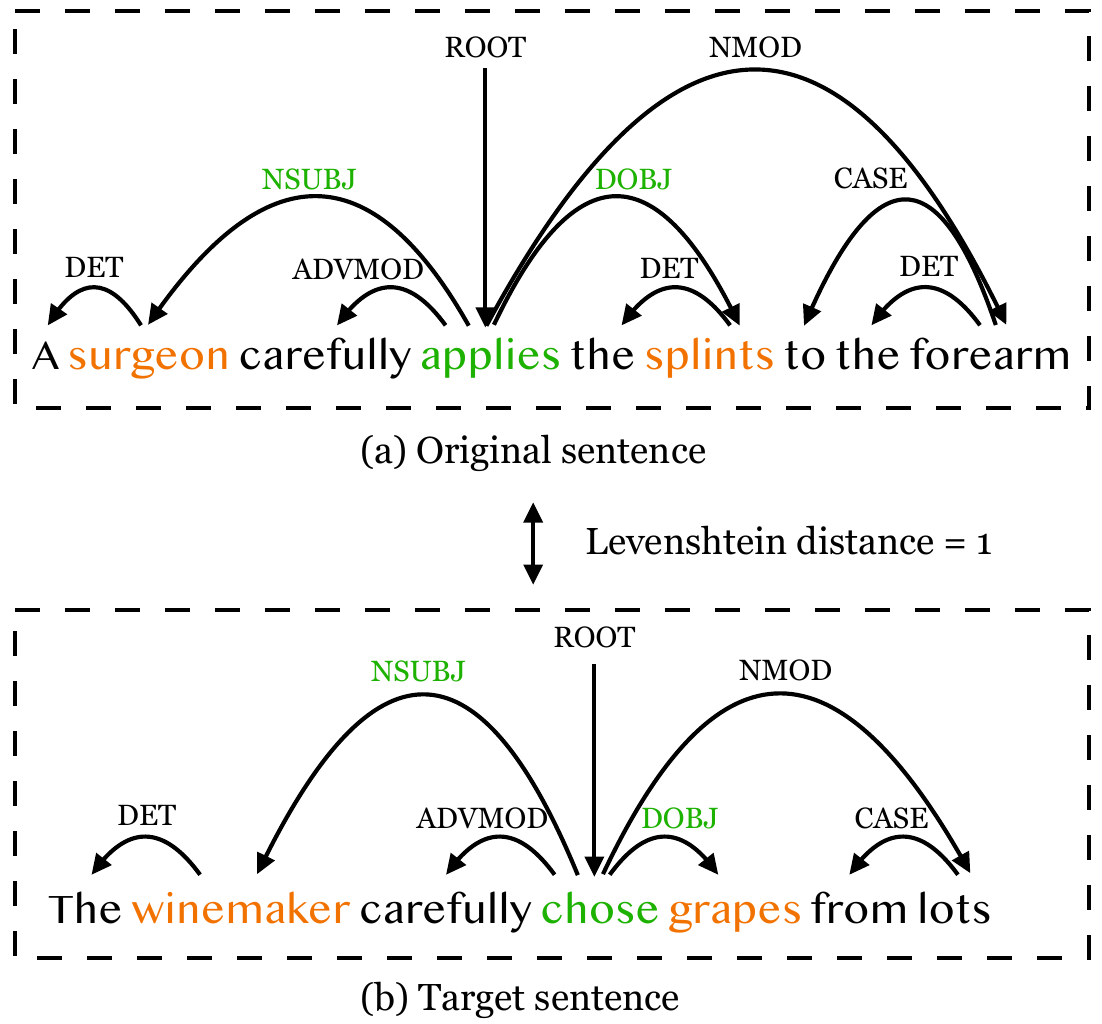}
    \caption{Overview of the original sentence approximation task. We highlight the \textcolor{orange}{entities} and \textcolor{gree}{pattern} in the original and target sentence.
    }
    \label{fig:distsnce}
    \vspace{-5mm}
\end{figure}

As illustrated in Figure \ref{fig:distsnce}, we first parse the original sentence \cite{chen2014fast} and obtain the path ``NSUBJ-applies-DOBJ'' between two entities ``surgeon'' and ``splints'' as pattern. To force the generative model to learn this pattern, we employ the Levenshtein (Lev) distance \cite{yujian2007normalized} to find the target pattern that is closest to the original pattern and has the same relation with the original sentence, then the corresponding sentence will be chosen as the output during training. 
% \chenwei{does the output include the pattern only or the whole sentence?}.
The Lev distance is a measure of the surface form similarity between two sequences, which is defined as the minimum number of operations (inserting, deleting, and replacing) required to convert the original sequence to the target sequence. We give a pseudo code implementation in the Appendix \ref{algorithm}.
For example, in Figure \ref{fig:distsnce}, the Lev distance between the pattern ``NSUBJ-applies-DOBJ'' and the pattern ``NSUBJ-chosen-DOBJ'' is $1$, so the target output sentence is:
\begin{figure}[h!]
\vspace{-2mm}
    \centering
    \includegraphics[width=0.99\linewidth]{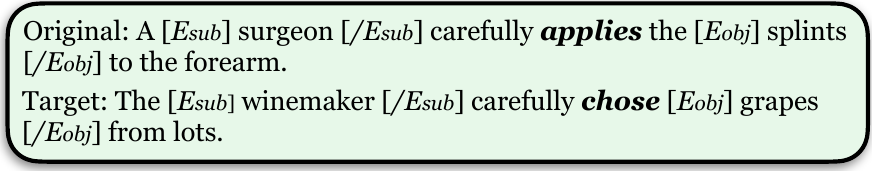}
\vspace{-9mm}
\end{figure}

In practice, we choose the target sentence with pattern less than $\lambda$ (e.g. $\lambda=3$) from the original sentence's pattern, where $\lambda$ is a hyperparameter.

In this way, the decoder network is employed to predict the pattern approximation target sentence ${Y = (y_{1}, y_{2}, ... , y_{T_{y}})}$ by maximizing ${p(Y|H,\theta_{P})}$:

\vspace{-2mm}
\begin{small}
\begin{equation}\label{Paa}
\mathcal{L}_{\theta_{P}}=\sum_{n=1}^{N}\sum_{i=1}^{T_{y}} \log p\left(y_{i}^{(n)} \mid y_{\textless i}^{(n)},H^{(n)}, \theta_{P}\right),
\end{equation}
\end{small}
% \vspace{-2mm}
where ${\theta_{P}}$ denotes the parameters of the decoder part for the original sentence pattern approximation. ${N}$ is the number of sentences for all outputs that satisfy the Lev distance less than $3$.

\paragraph{Entity-level sentence generation.}
Furthermore, to generate more controllable entity-level sentences and help the generator to better mark entities in the augmented sentences, we add a subject or object \textbf{Entity} ${(E)}$ from the target output sentence to the input embedding of the decoder as a hint.

For example, in Figure \ref{fig:overview}, we add 
% \chenwei{how we select this? always the head token, entity that appear first, or sampled randomly between head/tail entity?}
\textbf{winemaker} or \textbf{grapes} to the input of the decoder part ${\theta_{P}}$, which helps derive entity-oriented controllable sentence and increase the diversity of generated sentences by adopting different entity hints. Therefore, 
% \chenwei{say ``the final loss function we adopt'' or ``we finalize the loss function as''->}
we finalize the loss function of Eq. \ref{Paa} as:

\vspace{-4mm}
\begin{small}
\begin{equation}
\mathcal{L}_{\theta_{P}}=\sum_{n=1}^{N}\sum_{i=1}^{T_{y}} \log p\left(y_{i}^{(n)} \mid y_{\textless i}^{(n)},E^{(n)},H^{(n)}, \theta_{P}\right).
\end{equation}
\end{small}

The overall loss function of multi-task learning is the sum of log probabilities of original sentence restructuring and pattern approximation tasks:

\vspace{-4mm}
\begin{small}
\begin{equation}
\begin{aligned}
\mathcal{L}_{\theta}=& \sum_{n=1}^{N}\sum_{i=1}^{T_{y}} \log p\left(Y_{i}^{(n)} \mid Y_{\textless i}^{(n)},E^{(n)},H^{(n)}, \theta\right)\\
& + \sum_{m=1}^{M}\sum_{i=1}^{T_{x}} \log p\left(X_{i}^{'(m)} \mid X_{\textless i}^{'(m)},H^{(m)}, \theta\right),
\vspace{-1mm}
\end{aligned}
\end{equation}
\end{small}
where ${\theta=(\theta_{E},\theta_{R},\theta_{P})}$. ${\theta_{E}}$ is the parameter of encoder part. In practice, we adopt an iterative strategy to train two complementary tasks. For each iteration, we first optimize the ${(\theta_{E}, \theta_{R})}$ framework in the restructuring task for five epochs. The optimized ${\theta_{E}}$ will be employed as the initial ${\theta_{E}}$ of the pattern approximation task. Next, we optimize the ${(\theta_{E}, \theta_{P})}$ framework for five epochs in the pattern approximation task, and the updated ${\theta_{E}}$ will be used in the next iteration. 
% We believe that the pattern approximation task is more difficult than the restructuring task\chenwei{what are the evidences we collect to have/support this believe?}, so ${\theta_{P}}$ will be optimized for more epochs. 
Finally, ${\theta_{E}}$ and ${\theta_{P}}$ will be adopted to generate augmented sentences.

\subsection{Generate pseudo sentences}
\begin{table*}[t]
\centering

\scalebox{0.63}{
\begin{tabular}{lcccccccccccccccc}
\thickhline
\multicolumn{1}{c}{\multirow{2}{*}{Methods / Datasets}} & \multicolumn{1}{c}{\multirow{2}{*}{PLMs}} &\multicolumn{1}{c}{\multirow{2}{*}{Para.}} &\multicolumn{4}{c}{SemEval}&  \multicolumn{4}{c}{TACRED} &   \multicolumn{4}{c}{TACREV} & \multicolumn{1}{c}{\multirow{2}{*}{AVG.}}  & \multicolumn{1}{c}{\multirow{2}{*}{$\Delta$}} 
\\ \cmidrule(lr){4-7} \cmidrule(lr){8-11} \cmidrule(lr){12-15}
&&&10\% & 25\% & 50\% & 100\%& 10\% & 25\% & 50\% & 100\% &  10\% & 25\% & 50\% & 100\%  \\
\midrule 
\textbf{Base (SURE)}$\dagger$ & BART-Large & 406M & 77.2 & 81.5 & 83.9 & 86.3 & 67.9 & 70.4 & 71.9 & 73.3  & 72.3 & 75.1 & 77.4 & 79.2 &76.4  &--\\
\multicolumn{1}{l}{+EDA$\dagger$} & --& -- & 77.9  & 82.0  & 84.4 & 86.7 & 68.6 & 71.0 & 72.2 & 73.8  & 72.7 &  76.0 & 77.9 & 79.6 &76.9 & \color{red}0.5 $\uparrow$\\
\multicolumn{1}{l}{+Paraphrase Graph$\dagger$} &-- &--& 77.8 & 81.8 & 84.4 & 86.6 & 68.4 & 70.9 & 72.3 & 73.7  & 72.9& 75.7 & 77.7  & 79.4 &76.8 & \color{red}0.4 $\uparrow$\\ 
\multicolumn{1}{l}{+MixText$\dagger$} & BERT-Base & 110M & 78.6 & 82.6 & 85.0 & 87.2  & 69.0 & 71.7 & 72.9 & 74.1  & 73.6 & 76.4 & 78.4  &  80.0 & 77.5&  \color{red}1.1 $\uparrow$\\
\multicolumn{1}{l}{+Ex2$\dagger$} & T5-Base & 220M & \underline{79.1} & 83.0 & 85.5 & 87.5 &69.6 &\underline{72.1} &73.2 & 74.3 & \underline{74.2} &76.7 &\underline{78.8} & 80.3& \underline{77.8} & \color{red}\underline{1.4} $\uparrow$\\ 
\multicolumn{1}{l}{+BackMix$\ddagger$} & mBART-Base & 140M  & 78.7 & 82.5  & 85.2 & 87.2 & 69.2 & 71.8 & 72.8 & 74.0 & 73.9 & 76.3 & 78.2 &80.0 & 77.5  & \color{red}1.1 $\uparrow$\\
\multicolumn{1}{l}{+Soft DA$\dagger$} & BERT-Base & 110M & 78.5 &82.4 &85.1 &87.0  &68.9 &71.7 &72.8 &74.0  &73.5 &76.5 &78.4 &79.9 &77.4  & \color{red}1.0 $\uparrow$ \\
\multicolumn{1}{l}{+LAMBADA$\dagger$} & GPT-2 & 117M & 78.4 & 82.4 & 85.0 & 87.1 & 69.1& 71.6 & 72.9 & 73.8 &73.6 &76.5 &78.3 &79.8 &77.4  & \color{red}1.0 $\uparrow$\\
\multicolumn{1}{l}{+DARE$\ddagger$} & GPT-2 & 117M&78.7 &82.6 &85.3 &87.2  &69.2 &71.7 &72.9 &74.1   &73.8 &76.5 &78.4 & 80.1 & 77.5 & \color{red}1.1 $\uparrow$ \\
\multicolumn{1}{l}{+Text Gen$\ddagger$} & GPT-2-Medium & 345M & 79.0 & \underline{83.2} &\underline{85.7} & \underline{87.7}  &\underline{69.7} &71.9 &\underline{73.4} &\underline{74.4}  &\underline{74.2} &\underline{76.8} &78.6 &\underline{80.4} & \underline{77.8}&\color{red}\underline{1.4} $\uparrow$ \\

\cmidrule(lr){2-17}
\rowcolor{gray!15}
\multicolumn{1}{l}{+\textbf{\texttt{GDA} (T5-Base)}}& T5-Base & 220M  & \textbf{79.7} & \textbf{83.6} & \textbf{85.9} & \textbf{88.0}  &\textbf{70.4} &\textbf{72.6} &\textbf{73.8} &\textbf{74.9} &\textbf{74.8} & \textbf{77.2} & \textbf{79.1} & \textbf{80.8} & \textbf{78.3} & \textbf{\color{red}1.9 $\uparrow$} \\
\rowcolor{gray!15}
\multicolumn{1}{l}{\quad \textit{w/o Approximation}}& T5-Base & 220M  & 78.8 & 82.7 & 85.2 &87.4  &69.2 &71.9 &73.0 &74.5 &74.1 & 76.9 & 78.6 & 80.4 & 77.7 & -- \\
\rowcolor{gray!15}
\multicolumn{1}{l}{\quad \textit{w/o Restructuring}}& T5-Base & 220M & 79.1 & 82.9 & 85.4 & 87.5 & 69.4 & 71.9 & 73.2 & 74.6 & 74.4 & 77.0& 78.8 &80.5  & 77.9 & -- \\
\rowcolor{gray!15}
\multicolumn{1}{l}{\quad \textit{w/o Two Tasks}}& T5-Base & 220M  & 78.3 &82.3  & 84.7 & 86.9 & 68.7 & 71.2 & 72.5 & 74.0& 73.2 & 76.2&  78.1& 79.6 & 77.1 & -- \\

\cmidrule(lr){2-17}
\rowcolor{gray!10}
\multicolumn{1}{l}{+\textbf{\texttt{GDA} (BART-Base)}}& BART-Base & 140M & 79.2 & 83.2 & 85.6 & 87.8  &69.7 &72.3 &73.3 &74.5 &74.4 & 76.8 & 78.7 &80.5 & 78.0 & \color{red}1.6 $\uparrow$ \\
\rowcolor{gray!10}
\multicolumn{1}{l}{+\textbf{\texttt{GDA} (T5-Small)}}& T5-Small & 60M & 79.0 & 82.9 & 85.4 & 87.6  &69.5 &72.1 &72.9 &74.1 &74.1 &76.5 & 78.4 &  80.2  & 77.7&\color{red}1.3 $\uparrow$\\
\midrule 
\midrule

\textbf{Base (RE-DMP)}$\dagger$ &BERT-Large & 340M & 76.4 & 81.1  &  83.4 &  85.9 & 67.3 & 70.0 & 71.1 & 72.4   & 71.5 & 74.7 & 77.0 & 78.5 & 75.8 &--\\
\multicolumn{1}{l}{+EDA$\dagger$} & --& --&76.9 &81.5 &83.7 &86.1 &67.9 &70.4 & 71.5 &72.6  & 72.1 & 75.1 &77.3 &78.7 & 76.2 &\color{red}0.4 $\uparrow$\\
\multicolumn{1}{l}{+Paraphrase Graph$\dagger$}&--& -- &77.0 & 81.5 & 83.8 &86.2  & 68.0 &70.4 & 71.4& 72.7 & 72.1 &75.0 &77.2 &78.7 & 76.2 & \color{red}0.4 $\uparrow$\\
\multicolumn{1}{l}{+MixText$\dagger$} & BERT-Base & 110M & 77.8 & 82.0 & 84.0 & 86.5 & 68.6& 70.8& 71.9 & 73.1 &72.6 &75.6 & 77.7&79.2 & 76.7& \color{red}0.9 $\uparrow$\\
\multicolumn{1}{l}{+Ex2$\dagger$}& T5-Base & 220M  & 78.3 & \underline{82.6} &84.6  &87.0  & \underline{69.1} &71.4 &\underline{72.6} & 73.5 & 73.1 & \underline{76.2}& 78.0&79.5 &77.2  & \color{red}1.4 $ \uparrow$\\
\multicolumn{1}{l}{+BackMix$\ddagger$} & mBART-Base & 140M &77.9 & 82.2 & 84.3 & 86.6 & 68.5 & 71.0 & 71.9&73.2  &73.0  & 75.7& 77.6& 79.3& 76.8 & \color{red}1.0 $\uparrow$\\
\multicolumn{1}{l}{+Soft DA$\dagger$} & BERT-Base & 110M& 78.0 &82.1  &83.9  &86.4  & 68.3 & 70.9& 71.8&72.9  & 72.8 &75.6 &77.5 &79.1 & 76.6  &\color{red}0.8 $\uparrow$ \\
\multicolumn{1}{l}{+LAMBADA$\dagger$} & GPT-2 & 117M &77.7  &82.1  & 83.9 & 86.4 & 68.5& 70.7& 71.7 &73.0  &72.9 & 75.5& 77.6& 79.2 & 76.8& \color{red}1.0 $\uparrow$\\
\multicolumn{1}{l}{+DARE$\ddagger$} & GPT-2 & 117M& 77.8&82.3  &84.2  &86.5  & 68.7 &71.0 &72.0 & 73.1 &73.1  & 75.8&77.7 & 79.2& 76.8  &\color{red}1.0 $\uparrow$ \\
\multicolumn{1}{l}{+Text Gen$\ddagger$} & GPT-2-Medium & 345M &\underline{78.5} & \underline{82.6} &\underline{84.8}  &\underline{87.1}  & 69.0  &\underline{71.6} &72.5 & \underline{73.6} & \underline{73.3} & \underline{76.2}& \underline{78.1}&\underline{79.7} &\underline{77.3}  &\color{red}\underline{1.5} $\uparrow$ \\

\cmidrule(lr){2-17}
\rowcolor{gray!15}
\multicolumn{1}{l}{+\textbf{\texttt{GDA} (T5-Base)}}& T5-Base & 220M  &\textbf{79.0} & \textbf{83.1} & \textbf{85.4} & \textbf{87.8}  &\textbf{69.7} &\textbf{72.2}  & \textbf{73.1} & \textbf{74.1}  &\textbf{74.0} & \textbf{76.7} & \textbf{78.2} &\textbf{80.1} &\textbf{77.9} &\textbf{\color{red}2.1 $\uparrow$}\\
\rowcolor{gray!15}
\multicolumn{1}{l}{\quad \textit{w/o Approximation}}& T5-Base & 220M  & 78.0 & 82.4 & 84.6 & 87.0 & 68.9 & 71.4 & 72.2 & 73.6& 73.2 & 76.0&  78.1&79.4  &77.1  & -- \\
\rowcolor{gray!15}
\multicolumn{1}{l}{\quad \textit{w/o Restructuring}}& T5-Base & 220M  & 78.2 & 82.5 & 84.8 & 87.1 & 69.1 &71.5  & 72.4 &73.9 &73.5  & 76.2& 78.0 &79.7  & 77.2 & -- \\
\rowcolor{gray!15}
\multicolumn{1}{l}{\quad \textit{w/o Two Tasks}}& T5-Base & 220M  &77.3  &81.8  & 84.0 & 86.3 &68.3  & 70.7 & 71.7 & 72.9& 72.4 &75.3 & 77.5 & 79.0 & 76.4 & -- \\

\cmidrule(lr){2-17}
\rowcolor{gray!10}
\multicolumn{1}{l}{+\textbf{\texttt{GDA} (BART-Base)}} & BART-Base & 140M& 78.6& 82.5 & 85.0 & 87.4 &69.1 &71.7 &72.6 &73.4 &73.4 & 76.3 & 78.2 & 79.7 & 77.3 & \color{red}1.5 $\uparrow$ \\
\rowcolor{gray!10}
\multicolumn{1}{l}{+\textbf{\texttt{GDA} (T5-Small)}}& T5-Small & 60M & 78.2 & 82.0 & 84.6 & 86.8  & 68.5 & 71.2 & 72.0 &73.0 &73.1 & 75.9& 77.8 & 79.3 & 76.9 &\color{red}1.1 $\uparrow$ \\

\thickhline
\end{tabular}}
% \vspace{-2mm}
\caption{Average micro F1 results over three runs in three RE datasets. $\dagger$ means we rerun the open source code and adopt the given parameters, $\ddagger$ means we product the code with the given parameters. We \underline{underline} the best results among the baseline models. PLMs and Para. mean the pre-trained models and the corresponding parameters used.}
\label{tab:verification}
\vspace{-3mm}
\end{table*}
After we obtain the trained seq2seq generator T5 ${(\theta_{E},\theta_{P})}$, which focuses on the reconstruction of diverse, semantically consistent, and coherent relational signals. We leverage the generator to generate entity-oriented pseudo sentences as augmented data. In practice, we randomly select an annotated sentence ${X}$ and one marked subject or object entity ${E}$ 
% \chenwei{more details on the selection process. randomly?} 
under the relation label to which the ${X}$ belongs from the annotated data. Then we obtain the augmented sentence by ${(X,E,\theta_{E},\theta_{P})}$, where subject and object entities (one of them is ${E}$) have been marked during the generation process.

The augmented sentences have the same relation label as the original sentences and have enough diversity with different sentences and entity hints randomly sampled from the annotated data.

\section{Experiments}
We conduct extensive experiments on three public datasets and low-resource RE setting to show the effectiveness of \texttt{GDA} and give a detailed analysis to show its advantages.

\subsection{Base Models and Baseline Introduction}\label{baseline}
We adopt two SOTA base models: 

(1) \textbf{SURE} \cite{lu2022summarization} creates ways for converting sentences and relations that effectively fill the gap between the formulation of summarization and RE tasks. 

(2) \textbf{RE-DMP} \cite{tian2022improving} leverages syntactic information to improve relation extraction by training a syntax-induced encoder on auto-parsed data through dependency masking.

We adopt three types of baseline models.

(1) \textbf{Rule-Based Techniques}: \textbf{EDA} \cite{wei2019eda} adopts synonym replacement, random insertion, random swap, and random deletion to augment the original sentences. 
\textbf{Paraphrase Graph} \cite{chen2020finding} constructs a graph from the annotated sentences and creates augmented sentences by inferring labels from the original sentences using a transitivity property. 

(2) \textbf{Example Interpolation Techniques}: Inspired by Mixup \cite{zhang2018mixup}, \textbf{MixText} \cite{chen2020mixtext} and \textbf{Ex2} \cite{lee2021neural} aim to interpolate the embeddings and labels of two or more sentences. \textbf{BackMix} \cite{jin2022admix} proposes a back-translation based method which softly mixes the multilingual augmented samples.

(3) \textbf{Model-Based Techniques}:
\textbf{Soft DA} \cite{gao2019soft} replaces the one-hot representation of a word by a distribution over the vocabulary and calculates it based on contextual information. 
\textbf{LAMBADA} \cite{anaby2020not} and \textbf{DARE} \cite{papanikolaou2020dare} fine-tune multiple generative models for each relation type to generate augmentations. \textbf{Text Gen} \cite{bayer2022data} proposes a sophisticated generation-based method that generates augmented data by incorporating new
linguistic patterns.

\subsection{Datasets and Experimental Settings}
Three classical datasets are used to evaluate our technique: the SemEval 2010 Task 8 (\textbf{SemEval}) \cite{hendrickx2010semeval}, the TAC Relation Extraction Dataset (\textbf{TACRED}) \cite{zhang2017position}, and the revisited TAC Relation Extraction Dataset (\textbf{TACREV}) \cite{alt2020tacred}. SemEval is a classical benchmark dataset for relation extraction task which consists of 19 relation types, with 7199/800/1864 relation mentions in training/validation/test sets, respectively. 

TACRED is a large-scale crowd-source relation extraction benchmark dataset which is collected from all the prior TAC KBP shared tasks. TACREV found that the TACRED dataset contains about 6.62\% noisily-labeled relation mentions and relabeled the validation and test set. TACRED and TACREV consist of 42 relation types, with 75049/25763/18659 relation mentions in training/validation/test sets.
% SemEval is a classical benchmark dataset for relation extraction task which consists of 19 relation types, with 7199/800/1864 relation mentions in training/validation/test sets, respectively. TACRED is a large-scale crowd-source relation extraction benchmark dataset which is collected from all the prior TAC KBP shared tasks. TACREV found that the TACRED dataset contains about 6.62\% noisily-labeled relation mentions and relabeled the validation and test set. TACRED and TACREV consist of 42 relation types, with 75049/25763/18659 relation mentions in training/validation/test sets.
% \vspace{-0.5mm}
% \subsection{Experiments Settings}

We train the T5-base \cite{raffel2020exploring} with the initial parameter on the annotated data and utilize the T5 default tokenizer with max-length as 512 to preprocess data. We use AdamW with $5e{-5}$ learning rate to optimize cross-entropy loss. Both \texttt{GDA} and all baseline augmentation methods augment the annotated data by \textbf{3x} for fair comparison.
% Since MixText interpolates the embeddings and labels of sentences, we use the embeddings output by \texttt{CLS} token to represent the relational signals of sentences. 
For the low-resource RE setting, We randomly sample 10\%, 25\%, and 50\% of the training data and use them for all data augmentation techniques. All augmented techniques can only be trained and augmented on these sampled data.

% For the evaluation metrics, we choose F1 score as the main metric.
% \vspace{-1mm}
\subsection{Main Results}
% \vspace{-0.5mm}
Table \ref{tab:verification} shows the average micro F1 results over three runs in three RE datasets. All base models could gain F1 performance improvements from the augmented data when compared with the models that only adopt original training data, which demonstrates the effectiveness of data augmentation techniques in the RE task.
For three RE datasets, Text Gen is considered the previous SOTA data augmentation technique. The proposed \texttt{GDA} technique consistently outperforms all baseline data augmentation techniques in F1 performance (with student's T test $p<0.05$). More specifically, compared to the previous SOTA: Text Gen, \texttt{GDA} on average achieves 0.5\% higher F1 in SemEval, 0.5\% higher F1 in TACRED, and 0.4\% higher F1 in TACREV across various annotated data and base models.

Considering the low-resource relation extraction setting when annotated data are limited, e.g. 50\% for SemEval, TACRED and TACREV, \texttt{GDA} could achieve an average boost of 0.5\% F1 compared to Text Gen. When less labeled data is available, 10\% for SemEval, TACRED, and TACREV, the average F1 improvement is consistent, and surprisingly increased to 0.8\%. We attribute the consistent improvement of \texttt{GDA} to the diverse and semantically consistent generated sentences that are exploited: we bootstrap the relational signals of the augmented data via multi-task learning, which could help generate entity-oriented sentences for relation extraction tasks.

% \chenwei{why we interested in this study? parameter efficiency is what we want to claim here? or we study this mainly for a fair comparison?}

To demonstrate the impact of different pre-trained language models (PLMs) on the quality of augmented data, we present the PLMs adopted by \texttt{GDA} and baseline augmentation techniques and their corresponding parameters in Table \ref{tab:verification}. An exciting conclusion is that compared to Text Gen, although we use PLMs with fewer parameters (345M vs. 220M), our augmentation effect is still improved by an astonishing 0.6\% compared to Text Gen, and a new SOTA for the RE task has been achieved. 
% \chenwei{one potential nice aspect i feel we didn't highlight enough is that our model leverage the connections among training examples themselves for augmentation, not just doing augmentation based on individual examples one by one, so it is more like a ``structured'' agumentation. this also relates to the question of what are the good aspects to categorize baselines (e.g. whether understand sentence structures; whether leverage sentence correlations or just focusing on one sentece at a time)? -- this would further highlight modeling advantages of our model comparing with baselines, in addition to performance differences}
Even though we adopt T5-Small (60M) in \texttt{GDA}, which has fewer parameters than BERT-Base and GPT-2 ($\approx$ 110M), the augmented data can still bring competitive F1 improvement. More specifically, \texttt{GDA} (T5-Small) can achieve F1 improvement of 0.9\% and 1.1\% on SURE and RE-DMP, respectively, which illustrates the effectiveness of \texttt{GDA} for data augmentation in RE task.

% \subsection{Analysis and Discussion}
\subsection{Ablation Study}
% \noindent \textbf{Ablation Study}

We conduct an ablation study to show the effectiveness of different modules of \texttt{GDA} on test set. \texttt{GDA} \textit{w/o Restructuring} is the proposed technique without the decoder part ${\theta_{R}}$ and only uses the original sentence pattern approximation task to train the T5. \texttt{GDA} \textit{w/o Approximation} is the proposed technique without the decoder part ${\theta_{P}}$ and entity hint from the target sentence, and we use ${\theta_{R}}$ for generation during both training/inference.
% \chenwei{for both training/inference? and we use theta R for generation during both training/inference?}. 
\texttt{GDA} \textit{w/o Two Tasks} directly fine-tunes T5 on the training data, only requiring that the input sentence to be from the same relation as the target sentence.
\begin{table}[t!]
\centering
\resizebox{0.97\linewidth}{!}{
\begin{tabular}{lcccc}
\thickhline
Methods / Datasets & PLMs & SemEval & TACRED  & TACREV   \\
\midrule 
\multicolumn{1}{l}{SURE} & BART-Large &86.3 & 73.3 & 79.2 \\
\multicolumn{1}{l}{+EDA} & -- &  86.7 & 73.8 & 79.6\\
\multicolumn{1}{l}{+Paraphrase Graph}& -- & 86.6  &73.7&79.4 \\
% \multicolumn{1}{l}{+MixText} & 87.47 & 65.49 &74.89 \\
\multicolumn{1}{l}{+Ex2}& T5-Base & 87.5 &74.3 & 80.3\\
\multicolumn{1}{l}{+Text Gen}& GPT-2-Medium & \underline{87.7} &\underline{74.4} & \underline{80.4}\\
% \multicolumn{1}{l}{+Soft DA} & 87.53  & 65.72& 74.60\\
% \multicolumn{1}{l}{+LAMBADA} & 87.80 &65.74 &75.19 \\
% \multicolumn{1}{l}{+DARE} & 88.00  & 65.80 & 75.39\\

\rowcolor{gray!15}
\multicolumn{1}{l}{+Restructured} & -- & 87.0  & 74.0 & 79.8\\
\rowcolor{gray!15}
\multicolumn{1}{l}{+Pattern}  & --& 87.2  & 74.1 & 80.0\\
\rowcolor{gray!15}
\multicolumn{1}{l}
{+\textbf{\texttt{GDA}}} & T5-Base& \textbf{88.0}  &  \textbf{74.9}& \textbf{80.8} \\
\thickhline
\end{tabular}}
% \vspace{-2mm}
\caption{We adopt SURE as the base model and use 100\% training data over three datasets. We report F1 results on the test sets. ``Restructured'' and ``Pattern'' mean to directly use restructured original sentences and pattern approximation target sentences as augmentations.
}
\label{tab:generativeablation}
\vspace{-4mm}
\end{table}

A general conclusion from the ablation results in Table \ref{tab:verification} is that all modules contribute positively to \texttt{GDA}. More specifically, without multi-task learning framework, \texttt{GDA} \textit{w/o Two Tasks} brings 1.3\% less F1 performance averaged over three datasets. Similarly, compared with the restructuring task, the pattern approximation task can bring more average improvement in F1 performance (0.6\% vs. 0.8\%), which also means that we need to focus more on the pattern approximation task when training T5.

\subsection{Generative Model Ablation Study}
We additionally study the effect of removing the generative model on the augmentation effect, that is, we directly use restructured original sentences and pattern approximation target sentences as augmented sentences. 
From Table \ref{tab:generativeablation}, we found that directly using restructured sentences and pattern approximation sentences as augmented data results in a 1.3\% drop in F1 performance compared to \texttt{GDA}, which indicates the necessity of using T5-Base to train augmented sentences.
These two augmented sentences are also rule-based techniques. Compared with other rule-based data augmentation techniques (EDA and Paraphrase Graph), they can bring an average F1 improvement of 0.4\%, which additionally illustrates the effectiveness of our modification of the original sentences on the RE tasks.

\begin{figure}[t!]
    \centering
    \includegraphics[width=0.99\linewidth]{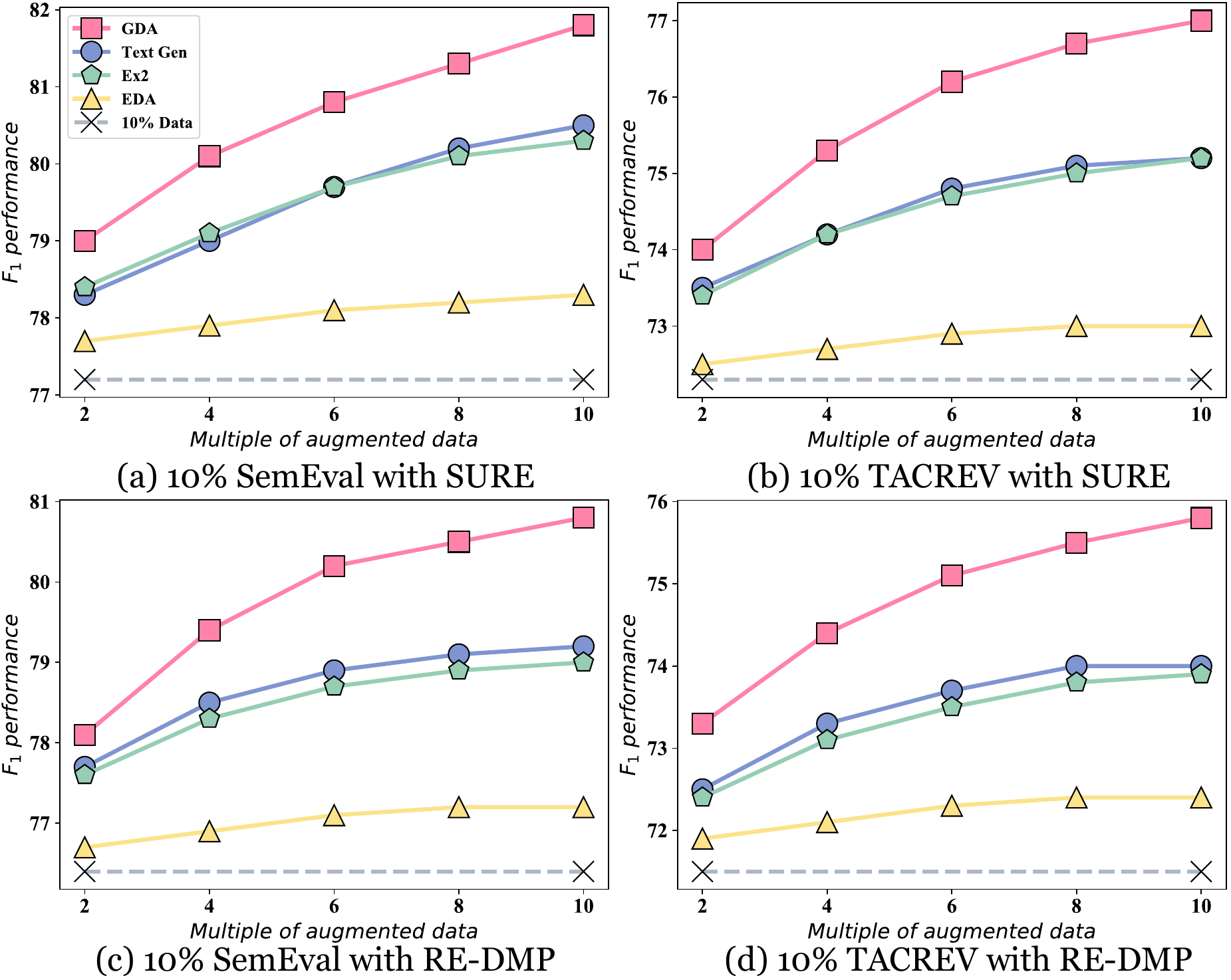}
    \caption{F1 results of the base model: SURE and RE-DMP with various multiples of the augmented data.}
    \label{fig:multiples}
\vspace{-4mm}
\end{figure}

\subsection{Performance on Various Augmentation Multiples}

We vary the multiple of augmented data from 2x to 10x the 10\% training set to study the influence of data augmentation techniques for the base models under low-resource scenarios. We choose the 10\% SemEval and 10\% TACREV training datasets and the base models: SURE and RE-DMP, then represent the results on the test set in Figure \ref{fig:multiples}.

We observe that two base models have more performance gains with ever-increasing augmented data and \texttt{GDA} achieves consistently better F1 performance, with a clear margin, when compared with baseline data augmentation techniques under various multiples of the augmented data. Especially for 10\% TACREV, \texttt{GDA} brings an incredible 3\% improvement in F1 performance with only 4x augmented data, which is even 0.2\% better than adopting 25\% (2.5x) of the training data directly.
% \vspace{-0.5mm}
\subsection{Diversity Evaluation}
We measure the diversity of augmented sentences through automatic and manual metrics. For automatic metric, we introduce the Type-Token Ratio (TTR) \cite{tweedie1998variable} to measure the ratio of the number of different words to the total number of words in the dependency path between two entities for each relation type. Higher TTR (\%) indicates more diversity in sentences. Besides that, we ask 5 annotators to give a score for the degree of diversity of the 30 generated sentences for each relation type, with score range of 1\textasciitilde5.
% we ask 5 annotators to label the relation of the 30 generated data for each relation type augmented by various augmentation techniques. According to the F1 performance of the relational labels, it can be judged whether the augmented data maintains the semantics of the original sentence. We also asked each annotator to give a score for the degree of \textbf{diversity} of the augmented sentences, with score range of 1\textasciitilde5. 
According to the annotation guideline in Appendix \ref{Guideline}, a higher score indicates the method can generate more diverse and grammatically correct sentences. 
We present the average scores for all relation types on three datasets in Table \ref{tab:human}. Since the example interpolation techniques do not generate the sentences shown, they are ignored. As a model-based augmentation technique, \texttt{GDA} could obtain 11.4\% TTR and 0.4 diversity performance boost in average compared to Text Gen, and can even have a diversity capability similar to the rule-based methods. Furthermore, we give the detailed hyperparameter analysis in Appendix.

\subsection{Case Study}\label{case}
We give two cases in Table \ref{tab:case_study}. \texttt{GDA} adopts the entity hint: ``program'' and input sentence to generate a controllable target sentence, while retaining the original pattern: ``was opened by'' without changing the semantics. \texttt{GDA} \textit{w/o Pattern Approximation} converts the rare pattern ``was opened by'' to the high frequency pattern ``consists of'' due to the inductive bias, which will affect the diversity of augmented sentences. \texttt{GDA} \textit{w/o Entity Hint} will generate uncontrollable entities, resulting in the same sentence generated by the same relation, which affects the diversity of generated sentences. 

\begin{table}[t!]
\centering
\resizebox{0.99\linewidth}{!}{
\begin{tabular}{l}
\thickhline
\begin{tabular}[c]{@{}l@{}}Original: The \textcolor{orange}{meeting} \textcolor{gree}{was opened by} the \textcolor{orange}{welcome speech} \\
${\quad\quad\quad}$ of the Mayor of Komotini.\\ 
{\color{purple}Relation Label}: \textbf{\texttt{Component-Whole}}\\
{\color{purple}Entity Hint}: \textbf{\texttt{program}}\\
\texttt{GDA}:The \textcolor{orange}{program} \textcolor{gree}{was opened by} the \textcolor{orange}{host}, who was a former \\
${\quad\quad\quad}$ member of the Congressional Black Caucus.\\
\texttt{GDA} \textit{w/o Pattern Approximation}: The \textcolor{orange}{program} \textcolor{gree}{consists of} a\\
${\quad\quad\quad}$ \textcolor{orange}{seminar} and a workshop.
\\ 
\texttt{GDA} \textit{w/o Entity Hint}: The ricotta \textcolor{orange}{mixture} \textcolor{gree}{was} the best \textcolor{gree}{part}\\ ${\quad\quad\quad}$ \textcolor{gree}{of} this \textcolor{orange}{dish}.

\end{tabular} 
\\
\hline
\begin{tabular}[c]{@{}l@{}}Original: This \textcolor{orange}{train} \textcolor{gree}{has} as many as six \textcolor{gree}{sets of} \textcolor{orange}{doors}.
 \\
{\color{purple}Relation Label}: \textbf{\texttt{Component-Whole}}\\
{\color{purple}Entity Hint}: \textbf{\texttt{kitchen}}\\
\texttt{GDA}: The \textcolor{orange}{kitchen} \textcolor{gree}{has sets of} \textcolor{orange}{stove}, and a microwave. \\
\texttt{GDA} \textit{w/o Pattern Approximation}: The \textcolor{orange}{kitchen} \textcolor{gree}{contents}
\\
${\quad\quad\quad}$ \textcolor{gree}{include} a \textcolor{orange}{refrigerator}.
\\ 
\texttt{GDA} \textit{w/o Entity Hint}: The ricotta \textcolor{orange}{mixture} \textcolor{gree}{was} the best \textcolor{gree}{part}\\ ${\quad\quad\quad}$ \textcolor{gree}{of} this \textcolor{orange}{dish}.

\end{tabular} 
\\
\thickhline
\end{tabular}
}
\caption{Case study. We highlight the \textcolor{orange}{entities} and \textcolor{gree}{pattern} in the original and generated sentences.}\label{tab:case_study}
% \vspace{-1mm}
\end{table}

\begin{table}[t]
\centering
\scalebox{0.65}{
\begin{tabular}{lcccccc}
\thickhline
\multicolumn{1}{c}{\multirow{2}{*}{Methods / Datasets}} & \multicolumn{2}{c}{SemEval} & \multicolumn{2}{c}{TACRED} &  \multicolumn{2}{c}{TACREV} 
\\ \cmidrule(lr){2-3} \cmidrule(lr){4-5} \cmidrule(lr){6-7}
&TTR & Diver.  &TTR  & Diver. &TTR & Diver.  \\
\midrule 
\multicolumn{1}{l}{EDA} & 82.4 & 3.1 & \textbf{84.7}  &  3.4 & 83.2 & 3.3  \\
\multicolumn{1}{l}{Paraphrase Graph} & \underline{85.9} &\underline{3.6}  & 84.1 & \underline{3.9} & \underline{84.6}  & 3.5 \\
\multicolumn{1}{l}{LAMBADA} & 72.6 & 2.3 & 76.2& 2.2 & 78.4 & 2.2\\
\multicolumn{1}{l}{DARE} & 75.3 & 2.4 &75.8 &  2.7 & 74.7 & 2.5 \\
\multicolumn{1}{l}{Text Gen} & 74.8 & 3.5 & 72.1 & 3.7 & 76.3 & \underline{3.6}\\
\rowcolor{gray!15}
\multicolumn{1}{l}{\texttt{GDA} (T5-Base)} &\textbf{86.4} &\textbf{4.0} & 84.1 & \textbf{4.1} &\textbf{86.9} & \textbf{3.9} \\
\rowcolor{gray!15}
\multicolumn{1}{l}{\texttt{GDA} \textit{w/o Approximation}} & 74.7 & 3.2 & 75.2 & 3.2 & 72.8 &3.1  \\
\rowcolor{gray!15}
\multicolumn{1}{l}{\texttt{GDA} \textit{w/o Restructuring}} & 80.3 & 3.8 & 81.3& 3.7  &82.0 & 3.8\\
% \hline
% \multicolumn{1}{l}{Original sentences} & 94.6 & 91.67 & 80.1 &  90.15 & 87.50 & 91.00\\

\thickhline
\end{tabular}}
% \vspace{-2mm}
\caption{Diversity Evaluation on three datasets.}
\label{tab:human}
\vspace{-3mm}
\end{table}

\subsection{Coherence Analysis}

Compared to rule-based augmentation techniques, \texttt{GDA} conditionally generates pseudo sentences with entity hints, providing more coherent and reasonable sentences. We analyze the coherence of the augmented sentences through perplexity based on GPT-2 \cite{radford2019language}. Note that the example interpolation techniques interpolate the embeddings and labels of two or more sentences without the generation of specific sentences, so we did not compare these methods. 
% We analyze the perplexity of the augmented sentences with various data augmentation techniques in three 100\% training datasets. 

From Table \ref{tab:perplexity}, \texttt{GDA} could obtain the lowest average perplexity.
% , even approximate the perplexity of the original sentences. 
Although Text Gen is also based on generative models, the augmented sentences are still not coherence enough due to the neglect of entity-level relational signals (entity hint) during the training process. Therefore, Text Gen is not so natural in generating augmented sentences with entity annotations.

\begin{table}[t!]
\centering
\resizebox{0.9\linewidth}{!}{
\begin{tabular}{lccc}
\thickhline
Methods / Datasets & SemEval & TACRED  & TACREV   \\
\midrule 
\multicolumn{1}{l}{EDA} &8.24  &9.18 &8.33 \\
\multicolumn{1}{l}{Paraphrase Graph}& 7.44  &7.88 &7.01 \\
\multicolumn{1}{l}{LAMBADA} & 4.21  &4.38 &\underline{4.11} \\
\multicolumn{1}{l}{DARE} & 4.28  &4.46 &4.22 \\
\multicolumn{1}{l}{Text Gen} & \underline{4.02}  &\underline{4.24} &\underline{4.11} \\
\rowcolor{gray!15}
\multicolumn{1}{l}{\texttt{GDA} (T5-Base)} & \textbf{3.97} &\textbf{4.21} &\textbf{4.05}  \\
\hline
\multicolumn{1}{l}{Original} & 3.88 &4.09 &3.91  \\
\thickhline
\end{tabular}}
% \vspace{-2mm}
\caption{Perplexity of the augmented sentences in three datasets. Original means the original sentences. Lower perplexity is better.}
\label{tab:perplexity}
\vspace{-4mm}
\end{table}

\subsection{Semantic Consistency Analysis}\

Unlike rule-based data augmentation techniques, which will change the semantics of the original sentence, \texttt{GDA} can better exploit relational signals: the target sentence during the training process comes from the restructured original sentence with the same relation label, so \texttt{GDA} can generate semantically consistent augmented sentences. 

\begin{figure}[t!]
    \centering
    \includegraphics[width=0.49\linewidth]{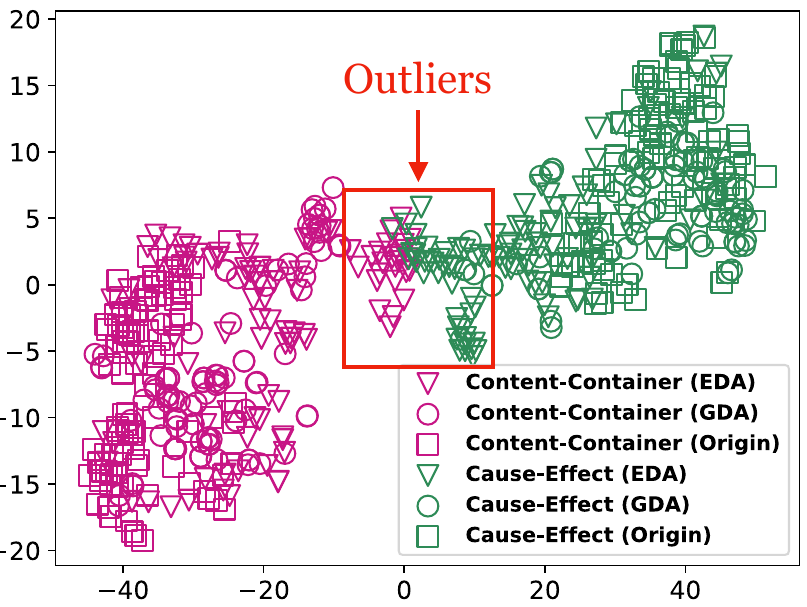}
    \includegraphics[width=0.49\linewidth]{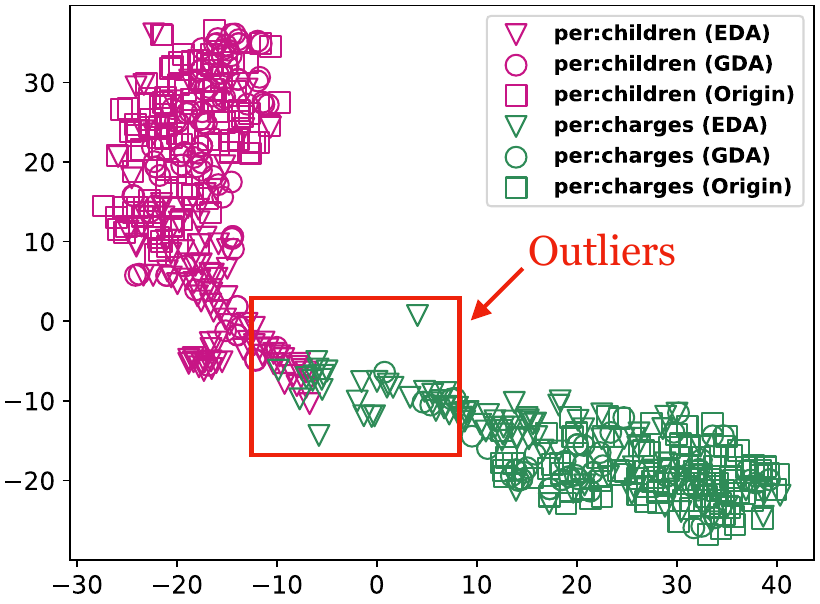}
    \caption{Latent space visualization of original and augmented sentences in the SemEval (left) and TACRED (right). The same relation labels use the same color.}
    % \chenwei{like the idea -- can we make it clearer in text or image? a bit hard to tell the distribution given each symbol. would it be helpful when we draw one figure for round+square and one for triangle+square (i feel that the main idea is to say/show rounds (not triangles) are better seperated among two colors)}}
    \label{fig:real}
\vspace{-4mm}
\end{figure}

In our tasks, we first train SURE on the 100\% training datasets and then apply \texttt{GDA} to the test set to obtain augmented sentences. We feed the 100 original sentences and 100 augmented sentences with the same relation labels into the trained SURE, and obtain the output representations from the last dense layer. We apply t-SNE \cite{van2014accelerating} to these embeddings and plot the visualization of the 2D latent space. From Figure \ref{fig:real}, we observed that the latent space representations of the augmented sentences closely surrounded those of the original sentences with the same relation labels, indicating that \texttt{GDA} could augment sentences semantically consistently. Conversely, sentences augmented with rule-based method: EDA appear outliers, indicating semantic changes.

% \subsection{Human Evaluation \chenwei{consider at least changing to ``diversity evaluation'', highlighting the objective not the annotation source}}

% \vspace{-0.5mm}
\section{Conclusions and Future works}
In this paper, we propose a relational text augmentation technique: \texttt{GDA} for RE tasks. Unlike conventional data augmentation techniques, our technique adopts the multi-task learning framework to generate diverse, coherent, and semantic consistent augmented sentences. We further adopt entity hints as prior knowledge for diverse generation.
% , which could help to generate entity-oriented sentences for RE tasks. 
% We find our data augmentation paradigm provides diverse, disambiguating, and generalizable signals to the relation extraction encoders.
Experiments on three public datasets and low-resource settings could show the effectiveness of \texttt{GDA}. For future research directions, we can try to explore more efficient pre-ordering and parsing methods, and apply our data augmentation methods to more NLP applications, such as semantic parsing \cite{liu2022semantic,liu2023comprehensive}, natural language inference \cite{li2023multi,li2022pair}.
% of \texttt{GDA} over competitive baselines.
\section{Limitations}
We would like to claim our limitations from two perspectives: application-wise and technical-wise. 

Application-wise: \texttt{GDA} needs annotations to fine-tune T5, which requires more computing resources and manual labeling costs than the rule-based techniques.

Technical-wise: Our ``original sentence restructuring'' and ``original sentence pattern approximation'' tasks rely on the efficiency and accuracy of pre-ordering rules \cite{wang2007chinese} and parsing methods \cite{chen2014fast}. Although current \texttt{GDA} show effectiveness, we still need to find more efficient pre-ordering and parsing methods. 

\section{Acknowledgement}
We thank the reviewers for their valuable comments. The work described here was partially supported by grants from the National Key Research and Development Program of China (No. 2018AAA0100204) and from the Research Grants Council of the Hong Kong Special Administrative Region, China (CUHK 14222922, RGC GRF, No. 2151185), NSF under grants III-1763325, III-1909323, III-2106758, and SaTC-1930941. 

% The work was supported by the National Key Research and Development Program of China (No. 2019YFB1704003), the National Nature Science Foundation of China (No. 62021002 and No. 71690231), NSF under grants III-1763325, III-1909323, III-2106758, SaTC-1930941, Tsinghua BNRist and Beijing Key Laboratory of Industrial Bigdata System and Application.
% \input{subfiles/7_ethics}

\bibliography{custom}
\bibliographystyle{acl_natbib}

% \clearpage
% \newpage
\appendix
\section{Levenshtein Distance Pseudo Code Implementation}\label{algorithm}
We give a pseudo code implementation of Levenshtein Distance algorithm.
\begin{lstlisting}[language=Go, float]
func LevenshteinDistance(char s[1..m], char t[1..n]){
  // for all i and j, d[i,j] will hold the Levenshtein distance between
  // the first i characters of s and the first j characters of t
  declare int d[0..m, 0..n]
 
  set each element in d to zero
 
  // source prefixes can be transformed into empty string by
  // dropping all characters
  for i from 1 to m:
    d[i, 0] := i
 
  // target prefixes can be reached from empty source prefix
  // by inserting every character
  for j from 1 to n:
    d[0, j] := j
 
  for j from 1 to n:
    for i from 1 to m:
      if s[i] = t[j]:
        substitutionCost := 0
      else:
        substitutionCost := 1

      d[i, j] := minimum(d[i-1, j] + 1,                   // deletion
                         d[i, j-1] + 1,                   // insertion
                         d[i-1, j-1] + substitutionCost)  // substitution
 
  return d[m, n]
}
\end{lstlisting}

% \section{Description of the Base Models}\label{base models}
% We adopt three classical base models and one SOTA base model to evaluate the quality of RE training datasets. (1) \textbf{PCNN} \cite{zeng2015distant}: Piecewise Convolutional Neural Network adopts CNN with piecewise max pooling to encode the sentence representation. Position embeddings are used to provide the encoder with the positions of subject and object entities in the sentence. (2) \textbf{PRNN} \cite{zhang2017position}: Integrating position-aware attention mechanism, the Position-aware Recurrent Neural Network evaluates the relative contribution of each word on the global positions after seeing the entire sentence. (3) \textbf{BERT} \cite{devlin2019bert}: Bidirectional Encoder Representations from Transformers is a multi-layer bidirectional Transformer based language representation model. It is designed to learn deep representations by jointly conditioning on the context of each word. In practice, we adopt BERT-base-uncased as the base model.

% \input{tables/PLM}

% \section{Parameters comparison of PLMs}\label{paraameter}
% We present the pre-trained language models (PLMs) adopted by \texttt{GDA} and baseline augmentation techniques and their corresponding parameters in Table \ref{tab:plm}.

\section{Hyperparameter Analysis}\label{Hyperparameter}

We study the hyperparameter $\lambda$ in the original sentence pattern approximation task, which represents the similarity between the pattern of the target sentence and the input sentence. An appropriate $\lambda$ can bring diverse generated sentences during generation process. We vary the $\lambda$ from 1 to 5 and represent the F1 results on the test set using the SURE model with 100\% training data in Table \ref{tab:hyperparameter}. 

With no more than 1\% F1 fluctuating results among three datasets, \texttt{GDA} is robust enough to $\lambda$. Besides, the results indicate that both approximation and coverage of target sentences will impact performance. Using a high $\lambda$ will introduce target sentences with low pattern approximation, causing the inductive bias problem. Low $\lambda$ will cause the low coverage of target sentences, affecting the training effect on T5.
\begin{table}[t!]
\centering
\resizebox{0.99\linewidth}{!}{
\begin{tabular}{lccccc}
\thickhline
Datasets / ${\lambda}$ & 1 & 2 & 3 & 4 & 5  \\
\midrule 
\multicolumn{1}{l}{SemEval} & 87.4 & 87.8  & \textbf{88.0} & 87.9 & 87.2 \\
\multicolumn{1}{l}{TACRED} & 73.8 &74.7 & \textbf{74.9} & 74.5 & 74.0\\
\multicolumn{1}{l}{TACREV} & 80.0 & 80.3 &80.8 & \textbf{81.0} & 80.4\\
% \multicolumn{1}{l}{Soft DA \cite{gao2019soft}}& 7.44  &7.88 &7.01 \\
\thickhline
\end{tabular}}
% \vspace{-2mm}
\caption{F1 performance under different ${\lambda}$.}
\label{tab:hyperparameter}
% \vspace{-2mm}
\end{table}

\section{Annotation Guideline}\label{Guideline}
Each annotator needs to carefully read each augmented sentence, compare it with the original sentence, and give a score according to the following criteria. Note that all augmented sentences for a relation label are given an average score, and the final score is the average of all relation labels.
\begin{itemize}
    \item Score:1. The augmented sentences under the same relation are almost the same.
    \item Score:2. The augmented sentences under the same relation are slightly different, with serious grammatical errors.
    \item Score:3. The augmented sentences under the same relation are slightly different, and there are almost no grammatical errors.
    \item Score:4. The augmented sentences under the same relation are diverse, with serious grammatical errors.
    \item Score:5. The augmented sentences under the same relation are diverse, and there are almost no grammatical errors.
\end{itemize}

\end{document}